\documentclass{article} 
\usepackage[T1]{fontenc}
\usepackage{iclr2027_conference,times}

\usepackage{amsmath,amsfonts,bm}

\def\eqref#1{equation~\ref{#1}}

\def\1{\bm{1}}

\DeclareMathAlphabet{\mathsfit}{\encodingdefault}{\sfdefault}{m}{sl}
\SetMathAlphabet{\mathsfit}{bold}{\encodingdefault}{\sfdefault}{bx}{n}

\newcommand{\softmax}{\mathrm{softmax}}

\usepackage[colorlinks,
            anchorcolor=red,
            citecolor=citecolor, 
            linkcolor=linkcolor,
            ]{hyperref}
\usepackage{url}

\usepackage{minitoc}
\usepackage[toc,page,header]{appendix}

\usepackage{algorithm}
\usepackage{algpseudocode} 
\usepackage{bm}
\usepackage{amsmath}
\usepackage{tikz}
\usetikzlibrary{arrows.meta}
\usepackage{tcolorbox}
\usepackage{amssymb}
\usepackage{threeparttable}
\usepackage{booktabs} 

\usepackage{wrapfig}
\usepackage{float}
\usepackage{hyperref}
\usepackage{xcolor}
\usepackage{graphicx}
\usepackage[table]{xcolor}
\usepackage[normalem]{ulem}

\definecolor{citecolor}{HTML}{0071BC}
\definecolor{linkcolor}{HTML}{ED1C24}
\definecolor{qwenpurple}{HTML}{800080}
\definecolor{reportblue}{HTML}{1F77B4}
\definecolor{jsonteal}{HTML}{008080}

\newcommand{\our}{DLoss}

\title{Diagnostic Loss: Asymmetric Supervision for Radiology Report Generation}

\iclrfinalcopy

\author{
\hspace*{-\tabcolsep}\parbox[t]{\textwidth}{\centering
\bfseries
Ignacy Kolton$^{1}$\quad
Wenxuan Li$^{2}$\quad
Zongwei Zhou$^{2}$\quad
Jaros\l{}aw B. \'{C}wik\l{}a$^{3}$ \\
Arkadiusz Sitek$^{4,5}$\quad
Pedro R. A. S. Bassi$^{2,*}$\quad
Szymon P\l{}otka$^{1,*}$\\
\normalfont\footnotesize
$^{1}$Jagiellonian University\quad
$^{2}$Johns Hopkins University\\
$^{3}$GAMMED\quad
$^{4}$Massachusetts General Hospital\quad
$^{5}$Harvard Medical School\\
\texttt{ignacy.kolton@doctoral.uj.edu.pl}\\[0.5em]
\normalsize*\footnotesize Joint co-supervision
}\hspace*{-\tabcolsep}
}

\begin{document}

\maketitle

\doparttoc 
\faketableofcontents 

\begin{abstract}
Automated radiology report generation requires models to do more than produce fluent clinical narratives: they must explicitly state supported diagnoses while avoiding unsupported ones.
However, existing medical vision-language models (MedVLMs) are typically trained with token-level maximum likelihood objectives,
which enforce general text agreement between AI-made and human-made radiology reports but do not prioritize diagnostically critical terms.
As a result, models
can generate reports that appear human-made but miss key and obvious diagnoses, and introduce wrong diagnoses.
We identify this as an objective misalignment between conventional language modeling and the asymmetric diagnostic requirements of radiology reporting.
To encode these requirements, we introduce \textbf{Diagnostic Loss (DLoss)}, a differentiable, architecture-agnostic objective that operates directly on decoder token probabilities.
DLoss separately aggregates probability mass over
vocabularies associated with diagnoses supported by the reference report and with unsupported alternatives.
Using asymmetric pooling across token positions in the report, it encourages supported diagnoses to receive sufficient probability in at least one position while suppressing spurious probability spikes for unsupported diagnoses
across all positions during training.
The loss can be applied to existing 3D MedVLMs without modifying their vision backbones, language decoders, or inference procedures, and introduces no additional inference-time computation.
We train and evaluate DLoss on 50,000 abdominal CT scans across several 3D MedVLM architectures on both in-distribution and out-of-distribution report-generation benchmarks.
Across these settings, DLoss improves diagnostic faithfulness while maintaining competitive performance on conventional language-generation metrics, showing the value of asymmetric sequence-level supervision for radiology report generation.
The code for DLoss is available at \url{https://github.com/gmum/DLoss}.
\end{abstract}

\section{Introduction}


Radiology reporting represents one of the most critical language-generation challenges in medicine \citep{boecking2022making}. Furthermore, reporting is essential across multiple medical imaging modalities, from single-frame 2D X-rays \citep{tiu2022expert, liu2024g2d, mh2026exgra, yang2026x} to 3D computed tomography (CT) volumes spanning hundreds of axial slices \citep{plotka2026mamba, zhou2026dual, rokuss2026voxtell}. The radiologist must collapse this dense visual evidence into concise text that states, in clinically explicit terms, which diagnoses are present or absent in the image \citep{jain1radgraph} (here, a diagnosis may refer to a disease or a radiological finding). The diagnostic value of a report lies precisely in this commitment: downstream clinical decisions, such as surveillance, biopsy, surgery, and oncologic treatment, are triggered by whether specific diagnostic terms appear and whether incorrect ones are omitted \citep{liu2024bootstrapping}. Automating elements of this process is highly appealing as imaging volumes continue to outpace radiologist capacity, but such automation is viable only if the generated report preserves the diagnostic value of the original report rather than merely mimicking its style and looking like a human-written report \citep{zhou2024benchx, wang2026curv,bassi2025radgpt}.

\begin{figure}[t]
    \centering
    \includegraphics[width=\linewidth]{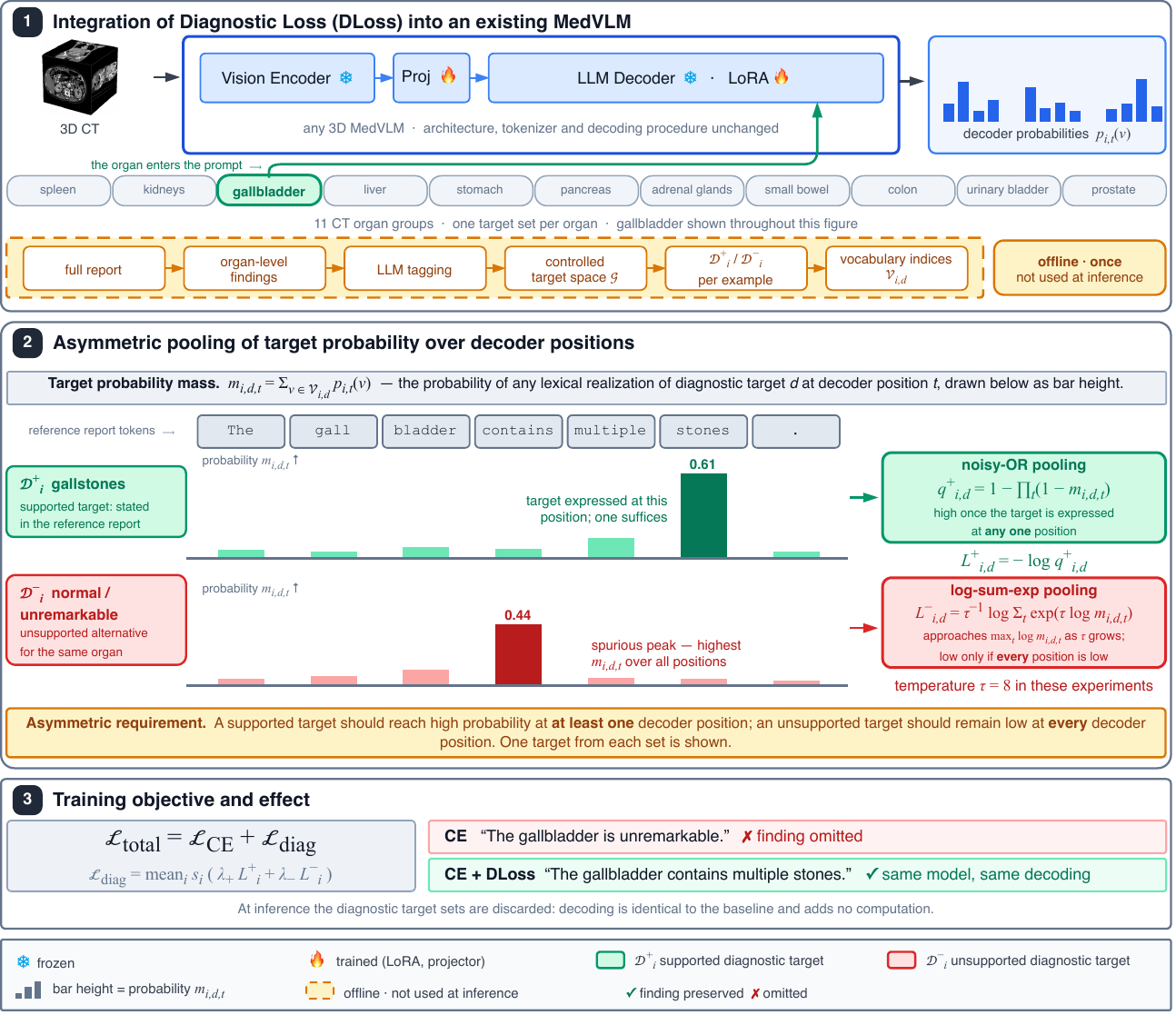}
    \caption{\textbf{Overview of \our{} and its workflow.} \our{} supervises how probability is distributed over diagnostic targets alongside standard token prediction. \textbf{(1)~Integration.} \our{} operates directly on decoder probabilities without modifying the architecture or decoding procedure (only the LoRA parameters and the projector are trained). Reports are decomposed into organ-level findings and mapped offline, via LLM tagging, to supported ($\mathcal{D}_i^{+}$) and unsupported ($\mathcal{D}_i^{-}$) diagnostic targets in a controlled target space $\mathcal{G}$. \textbf{(2)~Asymmetric pooling.} Target probability masses $m_{i,d,t}$ across valid decoder positions are aggregated using distinct operators: noisy-OR for supported targets (saturates when present at any position) and temperature-controlled log-sum-exp for unsupported targets (ensures probability stays low everywhere). \textbf{(3)~Objective.} The supported and unsupported target losses are added to cross-entropy with independent coefficients $\lambda_+$ and $\lambda_-$, respectively, and scaled by sample weight $s_i$. The diagnostic target sets are discarded at inference, so \our{} adds no inference-time computation.}
    
    \label{fig:diagnostic_loss}
\end{figure}

Recently, vision-language models (VLMs) trained on medical image-report pairs \citep{shuilarge, cao2025boosting, claessens2026scaling, zhao2026towards, hamamci2026better, hamamci2026generalist, blankemeier2026merlin} have brought this goal significantly closer.
By coupling visual encoders with large language models, modern medical VLMs (MedVLMs) produce fluent, well-structured reports and have been increasingly applied for report generation~\citep{shuilarge, cao2025boosting, kalisch2025ct, hamamci2026better,blankemeier2026merlin, yan2026learning}.
However, a recurring failure mode persists across architectures:
generated reports can look very similar to human-written reports while omitting diagnoses
that are present in the image, under-calling their severity, or
introducing diagnoses that the image does not support
\citep{delbrouck2022improving,yu2023evaluating,ostmeier2024green,chengrethinking,bassi2025radgpt}.
Radiologist audits of generated chest radiographs score these as clinically
significant errors, false diagnoses and omitted diagnoses, precisely when
they would alter clinical workflow \citep{yu2023evaluating,ostmeier2024green}.
In radiology and clinical oncology, missing or weakening an actionable
diagnosis is more consequential than a stylistic error, because subsequent
staging and treatment depend on whether the disease is named and on whether
unsupported differentials are introduced
\citep{bruno2015errors,bruno2017uncertainty}.

This mismatch between generating reports that look human-written and achieving diagnostic correctness
is not explained by modeling capacity alone; it also reflects a misalignment
in the training objective
\citep{miura2021improving,delbrouck2022improving,chengrethinking}.
Standard maximum-likelihood training optimizes token-level cross-entropy
against a ground-truth report \citep{jing2018automatic,delbrouck2022improving}.
This objective treats diagnostic and non-diagnostic tokens through the same
token-level loss: frequent narrative tokens dominate the gradient, rare
diagnostic terms are under-weighted, and there is no explicit sequence-level supervision encouraging supported diagnostic terms
to receive sufficient probability at some token position in the report
while suppressing probability peaks for unsupported terms across all positions
\citep{welleck2020neural,wu2023timer,liu2026satrrg}.
There is, therefore, little training-time pressure to satisfy the radiologist's
two-sided diagnostic obligation---supported diagnoses must be named, and
unsupported diagnoses must not be introduced
\citep{hartung2020create,yu2023evaluating,ostmeier2024green}.
Consequently, even strong MedVLMs can produce diagnostically incomplete or
misleading reports despite strong scores on conventional language metrics
\citep{papineni2002bleu,lin2004rouge,miura2021improving,delbrouck2022improving,yu2023evaluating}.

To address this misalignment in the training objective, we introduce Diagnostic Loss (DLoss), an auxiliary objective for radiology report generation that directly targets this two-sided diagnostic requirement: enforcing correct diagnoses and penalizing incorrect ones. DLoss is fully differentiable, operates directly on the decoder's token distribution, and is architecture-agnostic. It can be integrated on top of 3D autoregressive MedVLMs that expose decoder logits without altering the underlying backbone, tokenizer, or report-level supervision. Concretely, it utilizes asymmetric pooling of token probabilities over per-sample diagnostic vocabularies: probability mass on supported diagnostic targets is aggregated across token positions in the report to reward expressing the correct diagnosis somewhere, while probability mass on unsupported diagnostic targets is pooled with a tempered soft maximum to penalize any spurious spikes across these positions during training. This formulation encodes the asymmetric requirements of supported and unsupported diagnostic targets. Our main contributions are as follows:

\begin{enumerate}

\item \textbf{Asymmetric diagnostic supervision.} We introduce \our{}, which encodes different sequence-level constraints for supported and unsupported diagnostic concepts through asymmetric pooling of decoder token probabilities across token positions in the report. Supported concepts are encouraged to receive sufficient probability at least once, while unsupported concepts are penalized for probability spikes across these positions during training.

\item \textbf{Tokenizer-agnostic concept supervision.}
We represent diagnostic targets at the concept level and map their associated words and phrases (e.g., \textit{gallstones} and \textit{cholelithiasis}) to the token vocabulary of each decoder using its existing tokenizer. This provides token-level diagnostic supervision while allowing the same objective to operate across different tokenizers and MedVLM architectures without introducing a task-specific prediction head or modifying the inference procedure.

\item \textbf{Cross-model and out-of-distribution evaluation.} We train and evaluate \our{} on 50,000 abdominal CT scans across several 3D MedVLMs, decoder and tokenizer architectures, and both in- and out-of-distribution report-generation benchmarks. In-distribution experiments are conducted on abdominal CT scans (Merlin), while out-of-distribution performance is evaluated on a private, in-house abdominal CT dataset.

\end{enumerate}
Across six 3D MedVLMs (Merlin, fVLM, BTB3D, HLIP, RadFM, and SPECTRE), adding \our{} yields mean gains of 4.05 GREEN points and 3.82 RadGraph-F1 points on the Merlin test set, and 4.28 and 2.96 points, respectively, on the external test set (see Table~\ref{tab:3dradiologyreportsperformance}). Mean improvements across BLEU, ROUGE-L, METEOR, and CIDEr also show consistent positive gains across both sets.

\section{Related work}

\noindent
\textbf{MedVLMs for Radiology Report Generation.} Radiology report generation has evolved from specialized image-to-text architectures toward more general medical vision--language models. Early 2D approaches used CNN--RNN or hierarchical recurrent architectures with attention~\citep{wang2018tienet,jing2018automatic}, followed by Transformer-based generators such as R2Gen~\citep{chen2020generating}. More recent approaches leverage pretrained vision--language representations, such as BiomedCLIP~\citep{zhang2025multimodal}, and generative VLMs or large language models, including LLM-CXR~\citep{lee2024llm}, CMCRL~\citep{chen2025cross}, MLRG~\citep{liu2025enhanced}, and AOR~\citep{li2026aor}. For volumetric CT, recent work includes fVLM~\citep{shuilarge}, RadFM~\citep{wu2025towards}, HLIP~\citep{zhao2026towards}, BTB3D~\citep{hamamci2026better}, SPECTRE~\citep{claessens2026scaling}, and Merlin~\citep{blankemeier2026merlin}. DLoss is complementary to these architectural and representation-learning advances, modifying supervision over the decoder distribution rather than the underlying vision or language architecture.


\noindent
\textbf{Clinical Supervision and Training Objectives.} Clinical fidelity is not fully captured by lexical similarity, motivating measures such as RadGraph-based scores~\citep{jain1radgraph,delbrouck2022improving} and GREEN~\citep{ostmeier2024green}. RadGraph-F1 and GREEN measure clinical relevance by extracting structured entity-relation knowledge graphs and leveraging domain-tuned models to explicitly penalize diagnostic errors like false positives, false negatives, or anatomical misallocations that traditional lexical metrics ignore. Clinical information has also been incorporated into training through reinforcement learning, differentiable disease supervision, and semantic rewards~\citep{liu2019clinically,lovelace2020learning,delbrouck2022improving}, token-level reweighting or unlikelihood~\citep{wu2023timer,welleck2020neural}, disease- or anatomy-aware generation~\citep{jin2024promptmrg,chengrethinking}, and preference or reinforcement learning~\citep{liu2025rrg,zhou2026enhancing}.

Among approaches that modify the training objective, SAT-RRG uses an LLM to locate semantic errors and adjust training at the affected token positions~\citep{liu2026satrrg}. Unlikelihood training lowers the probability of selected negative candidates, including in sequence-level variants~\citep{welleck2020neural}. DLoss instead defines supported and unsupported diagnostic targets from each reference report and pools the decoder probabilities for each target across positions. Its two pooling operations encourage probability for a supported target without specifying where it must occur, while limiting probability spikes for unsupported targets throughout the training sequence. The pooling operations draw on multiple-instance learning~\citep{maron1997framework,wang2018comparing,pinheiro2015image}.

\section{Methodology}

\subsection{Preliminaries}

\our{} augments standard autoregressive report generation with
diagnostic supervision applied directly to the language decoder. The method
does not modify the visual encoder, language decoder, tokenizer, or inference
procedure. Instead, it operates on the decoder probabilities already computed
during report generation training.

\noindent
\textbf{Autoregressive Report Generation.} Consider a minibatch of $B$ radiology report generation examples. Each example $i$ is represented as $(x_i, y_i)$, where $x_i$ denotes the conditioning input,
including the image-derived representation and textual prompt, and $y_i = (y_{i,1}, \ldots, y_{i,T_i})$ denotes the reference report tokenized
over a vocabulary of size $V$. At output position $t$, the language decoder conditions on $x_i$ and the reference prefix $y_{i,<t}$ to produce a logit
vector $\bm{\ell}_{i,t} \in \mathbb{R}^{V}$. We denote the probability assigned to vocabulary item $v$ by:

\begin{equation}
    \label{eq:token-probability}
    p_{i,t}(v)
    =
    \softmax(\bm{\ell}_{i,t})_v .
\end{equation}

Let $\mathcal{P}_i$ denote the set of valid report-token positions for example
$i$, excluding prompt, padding, and image positions. The baseline training
objective is the token-level cross-entropy loss over the minibatch:

\begin{equation}
    \label{eq:cross-entropy-objective}
    \mathcal{L}_{\mathrm{CE}}
    =
    -\frac{1}{\sum_{i=1}^{B}|\mathcal{P}_i|}
    \sum_{i=1}^{B}
    \sum_{t \in \mathcal{P}_i}
    \log p_{i,t}(y_{i,t}) .
\end{equation}

Cross-entropy provides local supervision for each reference token, but does not
explicitly distinguish diagnostically relevant concepts from the remaining
report text. \our{} complements this objective with supervision over diagnostic
targets that should receive probability during decoding and diagnostic targets
that should remain unlikely.

\noindent
\textbf{Anatomy-level supervision for 3D CT.} For the 3D CT setting, we define diagnostic supervision at the anatomical
level, following prior anatomy-aware CT vision--language approaches
\citep{shuilarge,cao2025boosting}. Each study--report pair is decomposed into
anatomy-level image--text pairs:
\begin{equation}
(\mathbf{X}_n^{I},\,\mathbf{X}_n^{T}) \;\longmapsto\;
\bigl\{(\mathbf{X}_{n,c}^{I},\,\mathbf{X}_{n,c}^{T})\bigr\}_{c\in\mathcal{C}_n},
\end{equation}
where $\mathbf{X}_{n,c}^{I}$ and $\mathbf{X}_{n,c}^{T}$ denote the image
region and corresponding findings for anatomical organs $c$. Anatomical
structures are segmented using
TotalSegmentator~\citep{wasserthal2023totalsegmentator}, while reports are
parsed into organ-level findings using Qwen3-30B-A3B. This decomposition also sets the scope of the loss: supported and
unsupported targets are assigned within each organ-level pair and evaluated
only over its report-token positions. Applying all organ targets to one
full-report sequence could give opposing signals for the same word (e.g., in a
report describing a liver cyst and normal kidneys, \textit{cyst} would be
encouraged for the liver and penalized for the kidneys).


\noindent
\textbf{Diagnostic target construction.} Let $\mathcal{G}$ denote a controlled set of diagnostic targets. For each training example $i$, we construct two subsets: $\mathcal{D}_i^+ \subseteq \mathcal{G}$ contains diagnostic targets supported
by the reference report, whereas
$\mathcal{D}_i^- \subseteq \mathcal{G}$ contains unsupported diagnostic
targets. In the 3D setting, these sets are constructed within the anatomical
unit defined above. Reference-report expressions are mapped to concepts in $\mathcal{G}$:
supported concepts populate $\mathcal{D}_i^+$ and unsupported alternatives
populate $\mathcal{D}_i^-$. Explicitly negated concepts are excluded from
both sets to avoid conflicting lexical supervision.

A diagnostic target represents a clinical concept rather than a single surface
form. Each target
$d \in \mathcal{D}_i^+ \cup \mathcal{D}_i^-$ is therefore associated with a
tokenizer-specific set of vocabulary indices
$\mathcal{V}_{i,d} \subseteq \{1,\ldots,V\}$ corresponding to lexical
realizations of that concept. For example, lexical expressions such as
\textit{cholelithiasis}, \textit{gallstones}, and \textit{stone} can contribute
to the same diagnostic target. We construct these lexical sets offline,
starting from each normalized target name and adding recurrent words and short
phrases from organ-specific training reports assigned to that target through
LLM-based semantic tagging and synonym consolidation. We filter non-diagnostic
expressions; the resulting sets are grounded in the training corpus and do not
cover every possible paraphrase. These lexical realizations are tokenized
separately for the tokenizer used by each MedVLM, yielding the corresponding
sets $\mathcal{V}_{i,d}$. Appendix~\ref{sec:appendix_diagnostic_supervision}
shows this target construction with an organ-specific example. 

Importantly, \our{} only requires the decoder logits and the resulting sets
$\mathcal{D}_i^+$, $\mathcal{D}_i^-$, and $\mathcal{V}_{i,d}$. Consequently, the objective is independent of the architecture used to obtain the visual representation and applies to any MedVLM that exposes decoder logits. Since lexical
realizations are represented through tokenizer vocabulary indices, the
formulation provides token-level diagnostic supervision rather than exact
phrase-level matching.

\subsection{Diagnostic Loss}
\label{sec:diagnostic-loss-formulation}

Radiology reporting imposes asymmetric requirements on supported and
unsupported diagnostic targets. During training, a supported diagnostic target
should receive sufficient probability at at least one valid decoder position,
whereas an unsupported diagnostic target should remain unlikely across all
valid decoder positions. DLoss encodes these two requirements using different
pooling operators across decoder positions.

\noindent
\textbf{Diagnostic target probability mass.} For diagnostic target $d$, we first aggregate the probability mass assigned to its tokenizer-specific vocabulary indices at output position $t$:

\begin{equation}
    \label{eq:target-mass}
    m_{i,d,t}
    =
    \sum_{v \in \mathcal{V}_{i,d}}
    p_{i,t}(v) .
\end{equation}

The quantity $m_{i,d,t}$ therefore represents the total probability assigned
to lexical evidence associated with diagnostic target $d$ at a particular
decoder position.

\noindent
\textbf{Supported-Target Pooling.} For a supported target $d \in \mathcal{D}_i^+$, evidence at any valid decoder position can support the diagnosis. We aggregate the target probabilities
across positions using noisy-OR pooling:

\begin{equation}
    \label{eq:positive-occurrence-probability}
    q_{i,d}^{+}
    =
    1 -
    \prod_{t \in \mathcal{P}_i}
    \left(1-m_{i,d,t}\right) .
\end{equation}


The corresponding supported-target loss is:

\begin{equation}
    \label{eq:positive-target-loss}
    L_{i,d}^{+}
    =
    -\log q_{i,d}^{+} .
\end{equation}

\begin{wrapfigure}{r}{0.53\textwidth}
\vspace{-25pt}
\begin{minipage}{0.53\textwidth}
\begin{algorithm}[H]
\caption{\textbf{\our{} computation during decoder training.}}
\label{alg:diagnostic-loss}
\footnotesize
\begin{algorithmic}[1]

\Require Decoder logits $\bm{\ell}_{i,t}$ and valid positions $\mathcal{P}_i$
\Require Diagnostic target sets $\mathcal{D}_i^{+}$ and $\mathcal{D}_i^{-}$
\Require Vocabulary-index sets $\mathcal{V}_{i,d}$
\Require Target weights $w_{i,d}^{+}, w_{i,d}^{-}$ and sample weights $s_i$
\Require Hyperparameters $\lambda_+$, $\lambda_-$, and $\tau$
\Require Cross-entropy loss $\mathcal{L}_{\mathrm{CE}}$
\Ensure Total training loss $\mathcal{L}_{\mathrm{total}}$

\State $\mathcal{I}_{\mathrm{diag}}
\gets
\{i :
\mathcal{D}_i^{+}\cup\mathcal{D}_i^{-}\neq\varnothing\}$

\For{$i=1,\ldots,B$}

    \For{$t\in\mathcal{P}_i$}
        \State $p_{i,t}(v)
        \gets
        \softmax(\bm{\ell}_{i,t})_v$
    \EndFor

    \For{$d\in\mathcal{D}_i^{+}\cup\mathcal{D}_i^{-}$}
        \For{$t\in\mathcal{P}_i$}
            \State $m_{i,d,t}
            \gets
            \sum_{v\in\mathcal{V}_{i,d}}p_{i,t}(v)$
        \EndFor
    \EndFor

    \For{$d\in\mathcal{D}_i^{+}$}
        \State $q_{i,d}^{+}
        \gets
        1-\prod_{t\in\mathcal{P}_i}(1-m_{i,d,t})$
        \State $L_{i,d}^{+}
        \gets
        -\log q_{i,d}^{+}$
    \EndFor

    \For{$d\in\mathcal{D}_i^{-}$}
        \State $L_{i,d}^{-}
        \gets
        \tau^{-1}
        \log\!\left(
            \sum_{t\in\mathcal{P}_i}
            \exp(\tau\log m_{i,d,t})
        \right)$
    \EndFor

    \State Compute $W_i^{+}$ and $W_i^{-}$
    from $w_{i,d}^{+}$ and $w_{i,d}^{-}$

    \State Compute $L_i^{+}$ and $L_i^{-}$ as weighted means

    \State $\ell_i^{\mathrm{diag}}
    \gets
    s_i\left(\lambda_+L_i^{+}+\lambda_-L_i^{-}\right)$

\EndFor

\If{$|\mathcal{I}_{\mathrm{diag}}|>0$}
    \State $\mathcal{L}_{\mathrm{diag}}
    \gets
    \frac{1}{|\mathcal{I}_{\mathrm{diag}}|}
    \sum_{i\in\mathcal{I}_{\mathrm{diag}}}
    \ell_i^{\mathrm{diag}}$
\Else
    \State $\mathcal{L}_{\mathrm{diag}}\gets 0$
\EndIf

\State $\mathcal{L}_{\mathrm{total}}
\gets
\mathcal{L}_{\mathrm{CE}}
+
\mathcal{L}_{\mathrm{diag}}$

\State \Return $\mathcal{L}_{\mathrm{total}}$

\end{algorithmic}
\end{algorithm}
\end{minipage}
\vspace{-60pt}
\end{wrapfigure}

The pooled probability increases when diagnostic evidence receives probability
at one or more valid decoder positions. Minimizing $L_{i,d}^{+}$ therefore
encourages the decoder to assign sufficient probability to a lexical
realization of the supported target without prescribing a particular output
position.

\noindent
\textbf{Unsupported-Target Pooling.} For an unsupported target $d \in \mathcal{D}_i^-$, a high probability at even a single position can introduce diagnostically inconsistent content. We
therefore aggregate target probabilities using a temperature-controlled
log-sum-exp over their logarithms:

\begin{equation}
    \label{eq:negative-target-loss}
    L_{i,d}^{-}
    =
    \tau^{-1}
    \log
    \left(
        \sum_{t \in \mathcal{P}_i}
        \exp\left(
            \tau \log m_{i,d,t}
        \right)
    \right) ,
\end{equation}

where $\tau > 0$ controls the sharpness of the aggregation. As $\tau$
increases, the pooling increasingly concentrates on the output position with
the largest unsupported target probability. Minimizing $L_{i,d}^{-}$ therefore
suppresses spurious probability peaks for unsupported diagnostic targets across
valid decoder positions. For numerical stability, arguments to logarithms are
clamped away from zero in the implementation; this stabilization is omitted
from the equations for readability.

\noindent
\textbf{Overall Training Objective.}
Let $w_{i,d}^{+}$ and $w_{i,d}^{-}$ denote non-negative target-level weights,
set to one in all reported experiments, and define
$W_i^{+} = \sum_{d \in \mathcal{D}_i^+} w_{i,d}^{+}$ and
$W_i^{-} = \sum_{d \in \mathcal{D}_i^-} w_{i,d}^{-}$.
The supported and unsupported diagnostic components for example $i$ are
computed as weighted means:

\begin{equation}
    \label{eq:sample-diagnostic-components}
    L_i^{\pm}
    =
    \begin{cases}
    \dfrac{
        \sum_{d \in \mathcal{D}_i^{\pm}}
        w_{i,d}^{\pm} L_{i,d}^{\pm}
    }{
        W_i^{\pm}
    },
    & W_i^{\pm} > 0, \\[0.8em]
    0,
    & W_i^{\pm} = 0 .
    \end{cases}
\end{equation}

The sample-level diagnostic loss is then:

\begin{equation}
    \label{eq:sample-diagnostic-loss}
    \ell_i^{\mathrm{diag}}
    =
    s_i
    \left(
        \lambda_+L_i^+
        +
        \lambda_-L_i^-
    \right) ,
\end{equation}

where $s_i \geq 0$ denotes the sample weight, and the non-negative coefficients
$\lambda_+$ and $\lambda_-$ independently control the contributions of
supported and unsupported diagnostic targets. Let
$\mathcal{I}_{\mathrm{diag}}
=
\{i \in \{1,\ldots,B\} :
\mathcal{D}_i^+ \cup \mathcal{D}_i^- \neq \varnothing\}$
denote the set of training examples containing at least one active diagnostic
target. The batch-level \our{} is defined as:


\begin{equation}
    \label{eq:diagnostic-loss}
    \mathcal{L}_{\mathrm{diag}}
    =
    \begin{cases}
    \dfrac{1}{|\mathcal{I}_{\mathrm{diag}}|}
    \displaystyle\sum_{i \in \mathcal{I}_{\mathrm{diag}}}
    \ell_i^{\mathrm{diag}},
    & |\mathcal{I}_{\mathrm{diag}}| > 0, \\[0.8em]
    0,
    & |\mathcal{I}_{\mathrm{diag}}| = 0 .
    \end{cases}
\end{equation}

Finally, \our{} is optimized jointly with the standard language-modeling objective: $\mathcal{L}_{\mathrm{total}} = \mathcal{L}_{\mathrm{CE}} + \mathcal{L}_{\mathrm{diag}}$,
where $\mathcal{L}_{\mathrm{diag}}$ contains the weighted supported- and
unsupported-target components defined above. Algorithm~\ref{alg:diagnostic-loss}
summarizes the computation of the training objective. \our{} is used only during training. At
inference, the diagnostic target sets are discarded and each MedVLM follows the same decoding procedure as its corresponding baseline. The proposed
objective therefore introduces no additional inference-time computation and
requires no modification of the underlying vision or language architecture.



\section{Experiments}

\subsection{Exprimental setup}

\noindent
\textbf{Datasets.} For training and evaluation, we employ three distinct 3D abdominal CT datasets paired with their corresponding radiology reports. Specifically, models are trained on CancerVerse~\citep{li2026cancerverse} (24,168 scans from 13,863 patients) and the training split of the publicly available Merlin~\citep{blankemeier2026merlin, bassi2026merlin} dataset (20,369 scans). Together, Merlin and CancerVerse represent the largest public abdominal CT datasets, containing over 20 million image slices. Evaluation is conducted on Merlin's official test set (5,125 scans) for in-distribution assessment, alongside an external, in-house dataset (2,706 scans from 1,948 patients) for out-of-distribution evaluation. Comprehensive details for all datasets are provided in Appendix~\ref{sec:appendixdata}.

\noindent
\textbf{Implementation details.} Pooled training CT volumes are provided as a converted 1.5-mm isotropic view. Each backbone then applies model-specific intensity normalization, geometric preprocessing, and feature extraction based on its original implementation. Scans from Merlin's official test split are supplied at native spacing and processed by the corresponding backbone-specific pipeline. Within each backbone, the CE and DLoss runs use identical CT-derived inputs. Training uses up to four NVIDIA GH200 GPUs, each equipped with 96 GB of VRAM. The implementation details for all experiments can be found in Appendix~\ref{sec:appendix_implementation_details}.

\noindent
\textbf{Evaluation metrics.} We use GREEN \citep{ostmeier2024green}, BLEU \citep{papineni2002bleu}, ROUGE-L \citep{lin2004rouge}, METEOR \citep{lavie2009meteor}, CIDEr \citep{vedantam2015cider}, and RadGraph-F1 \citep{jain1radgraph} to evaluate the performance of report generation. While metrics like BLEU, ROUGE-L, METEOR, and CIDEr measure surface-level n-gram overlap, RadGraph-F1 and GREEN evaluate diagnostic accuracy. RadGraph-F1 extracts anatomical entities, findings, and their relationships into knowledge graphs to verify factual consistency. GREEN uses a domain-adapted language model to explicitly identify and weight clinically critical errors, such as false positives, false negatives, or anatomical mislocations, aligning its score directly with radiologist judgment rather than phrasing. A detailed description can be found in Appendix~\ref{sec:appendix_evaluation_metrics}. 

\noindent
\textbf{Enhancing state-of-the-art baselines.} To demonstrate the broad utility and fair contribution of our proposed \our{}, we integrate it across a diverse set of leading MedVLMs, including Merlin \citep{blankemeier2026merlin}, SPECTRE \citep{claessens2026scaling}, RadFM \citep{wu2025towards}, fVLM \citep{shuilarge}, BTB3D \citep{hamamci2026better}, and HLIP \citep{zhao2026towards}. Crucially, we evaluate each model fine-tuned with vs. without \our{} under identical experimental conditions, directly isolating the performance gains attributable to our diagnostic objective from standard supervised fine-tuning.


\subsection{Experimental results}

\textbf{Quantitative results.} Table~\ref{tab:3dradiologyreportsperformance} shows that adding \our{} improves report generation across six 3D MedVLMs on both the Merlin and external test sets. Averaged over backbones, GREEN and F1-RG increase by 4.05 and 3.82 points on Merlin, and by 4.28 and 2.96 points externally. Mean gains are also positive for BLEU, ROUGE-L, METEOR, and CIDEr on both sets. Overall, 71 of the 72 model--metric--test comparisons improve, including every comparison on the external set. The sole exception is a 0.01-point decrease in F1-RG for the Merlin backbone. A complementary nine-seed analysis of training variability and bootstrap uncertainty is provided in Appendix~\ref{sec:appendix_nine_seed_intervals}.

\begin{table}[t]
\centering
\caption{\textbf{Quantitative evaluation of 3D radiology report generation on internal and external benchmarks.} We report language generation and clinical accuracy metrics on the Merlin (\textbf{M}) test set and an external (\textbf{E}) in-house dataset. F1-RG denotes the complete RadGraph score. Rows labeled \textbf{+~\our{}} reflect signed changes relative to the baseline model after training with \our{}. Gains are shown in \textcolor{green!55!black}{green} and drops in \textcolor{red!90!black}{red}. $\bar{\Delta}$ represents the mean change over the corresponding baseline.}
\label{tab:3dradiologyreportsperformance}
\begin{threeparttable}
\small
\setlength{\tabcolsep}{3.5pt}
\resizebox{\linewidth}{!}{%
\begin{tabular}{lcccccccccccc}
\toprule[1pt]
Method
& \multicolumn{2}{c}{BLEU $\uparrow$}
& \multicolumn{2}{c}{ROUGE-L $\uparrow$}
& \multicolumn{2}{c}{METEOR $\uparrow$}
& \multicolumn{2}{c}{CIDEr $\uparrow$}
& \multicolumn{2}{c}{GREEN $\uparrow$}
& \multicolumn{2}{c}{F1-RG $\uparrow$}
\\
\cmidrule(lr){2-3}
\cmidrule(lr){4-5}
\cmidrule(lr){6-7}
\cmidrule(lr){8-9}
\cmidrule(lr){10-11}
\cmidrule(lr){12-13}
& M & E
& M & E
& M & E
& M & E
& M & E
& M & E
\\
\midrule[0.5pt]
Merlin
& 20.65 & 16.63
& 44.21 & 34.30
& 18.45 & 15.21
& 13.98 & 10.89
& 53.82 & 45.01
& 35.01 & 26.36
\\
\hspace{0.5em} + \our{}
& \textcolor{green!55!black}{+0.01} & \textcolor{green!55!black}{+0.09}
& \textcolor{green!55!black}{+0.05} & \textcolor{green!55!black}{+0.26}
& \textcolor{green!55!black}{+0.01} & \textcolor{green!55!black}{+0.07}
& \textcolor{green!55!black}{+0.18} & \textcolor{green!55!black}{+0.50}
& \textcolor{green!55!black}{+0.83} & \textcolor{green!55!black}{+1.24}
& \textcolor{red!90!black}{$-$0.01} & \textcolor{green!55!black}{+0.35}
\\
fVLM
& 39.58 & 18.61
& 50.02 & 34.06
& 26.32 & 15.75
& 36.25 & 18.15
& 53.91 & 46.78
& 46.53 & 29.32
\\
\hspace{0.5em} + \our{}
& \textcolor{green!55!black}{+7.04} & \textcolor{green!55!black}{+2.59}
& \textcolor{green!55!black}{+6.70} & \textcolor{green!55!black}{+2.39}
& \textcolor{green!55!black}{+4.04} & \textcolor{green!55!black}{+1.26}
& \textcolor{green!55!black}{+6.56} & \textcolor{green!55!black}{+1.34}
& \textcolor{green!55!black}{+8.99} & \textcolor{green!55!black}{+3.06}
& \textcolor{green!55!black}{+6.56} & \textcolor{green!55!black}{+1.65}
\\
BTB3D
& 15.96 & 16.38
& 32.21 & 33.75
& 17.07 & 16.14
& 17.39 & 19.50
& 53.38 & 55.76
& 25.82 & 27.95
\\
\hspace{0.5em} + \our{}
& \textcolor{green!55!black}{+8.62} & \textcolor{green!55!black}{+6.74}
& \textcolor{green!55!black}{+7.47} & \textcolor{green!55!black}{+4.39}
& \textcolor{green!55!black}{+3.99} & \textcolor{green!55!black}{+2.42}
& \textcolor{green!55!black}{+10.52} & \textcolor{green!55!black}{+4.32}
& \textcolor{green!55!black}{+2.05} & \textcolor{green!55!black}{+2.06}
& \textcolor{green!55!black}{+9.18} & \textcolor{green!55!black}{+4.95}
\\
HLIP
& 40.18 & 21.54
& 52.16 & 40.70
& 26.82 & 18.07
& 37.33 & 21.73
& 56.15 & 49.54
& 48.02 & 34.22
\\
\hspace{0.5em} + \our{}
& \textcolor{green!55!black}{+2.44} & \textcolor{green!55!black}{+2.98}
& \textcolor{green!55!black}{+2.02} & \textcolor{green!55!black}{+3.47}
& \textcolor{green!55!black}{+1.51} & \textcolor{green!55!black}{+1.61}
& \textcolor{green!55!black}{+1.74} & \textcolor{green!55!black}{+2.40}
& \textcolor{green!55!black}{+4.09} & \textcolor{green!55!black}{+4.43}
& \textcolor{green!55!black}{+2.18} & \textcolor{green!55!black}{+3.38}
\\
RadFM
& 49.93 & 27.80
& 60.55 & 45.09
& 32.84 & 21.06
& 48.04 & 27.41
& 64.80 & 62.71
& 57.91 & 39.74
\\
\hspace{0.5em} + \our{}
& \textcolor{green!55!black}{+0.06} & \textcolor{green!55!black}{+0.21}
& \textcolor{green!55!black}{+0.22} & \textcolor{green!55!black}{+0.48}
& \textcolor{green!55!black}{+0.06} & \textcolor{green!55!black}{+0.18}
& \textcolor{green!55!black}{+0.11} & \textcolor{green!55!black}{+0.25}
& \textcolor{green!55!black}{+0.50} & \textcolor{green!55!black}{+0.86}
& \textcolor{green!55!black}{+0.23} & \textcolor{green!55!black}{+0.41}
\\
SPECTRE
& 31.88 & 14.20
& 42.54 & 28.14
& 22.89 & 12.64
& 32.30 & 13.58
& 45.35 & 41.34
& 41.95 & 26.53
\\
\hspace{0.5em} + \our{}
& \textcolor{green!55!black}{+5.11} & \textcolor{green!55!black}{+10.86}
& \textcolor{green!55!black}{+5.41} & \textcolor{green!55!black}{+10.95}
& \textcolor{green!55!black}{+2.79} & \textcolor{green!55!black}{+6.59}
& \textcolor{green!55!black}{+3.01} & \textcolor{green!55!black}{+10.12}
& \textcolor{green!55!black}{+7.84} & \textcolor{green!55!black}{+14.03}
& \textcolor{green!55!black}{+4.79} & \textcolor{green!55!black}{+7.03}
\\
\midrule
{$\bar{\Delta}$}
& \textbf{\textcolor{green!55!black}{+3.88}} & \textbf{\textcolor{green!55!black}{+3.91}}
& \textbf{\textcolor{green!55!black}{+3.65}} & \textbf{\textcolor{green!55!black}{+3.66}}
& \textbf{\textcolor{green!55!black}{+2.07}} & \textbf{\textcolor{green!55!black}{+2.02}}
& \textbf{\textcolor{green!55!black}{+3.69}} & \textbf{\textcolor{green!55!black}{+3.16}}
& \textbf{\textcolor{green!55!black}{+4.05}} & \textbf{\textcolor{green!55!black}{+4.28}}
& \textbf{\textcolor{green!55!black}{+3.82}} & \textbf{\textcolor{green!55!black}{+2.96}}
\\
\bottomrule[1pt]
\end{tabular}%
}
\end{threeparttable}
\end{table}


\noindent
\textbf{Computational efficiency.} Table~\ref{tab:computational_cost} compares CE training with CE + \our{} across four 3D MedVLMs. The largest measured mean increase in step time is 4.4\% (BTB3D); peak allocated and reserved memory increase by at most 1.6\% and 2.9\%, respectively. Since \our{} is used only during training, it adds no inference-time computation.

\begin{table}[bp]
\centering
\caption{\textbf{Measured training-step cost of CE and CE plus \our{}.}
Profiling used matched batches of eight in BF16 on a single NVIDIA GH200 GPU. Step times average 12 fresh-process runs per condition across six counterbalanced blocks; time $\Delta$ is the mean paired relative change across blocks. Peak allocated and reserved memory average three separate runs per condition, with $\Delta$ reporting relative changes between condition means. Allocated memory measures live tensors, while reserved memory includes allocator-held capacity. The optimized \our{} implementations used for profiling were validated against the training objective to ensure identical loss values and gradients.}

\label{tab:computational_cost}
\small
\setlength{\tabcolsep}{3pt}
\begin{tabular*}{\linewidth}{@{\extracolsep{\fill}}lrrrrrrrrr@{}}
\toprule[1pt]
Method
& \multicolumn{3}{c}{Step time (ms)}
& \multicolumn{3}{c}{Peak allocated (GiB)}
& \multicolumn{3}{c}{Peak reserved (GiB)} \\
\cmidrule(lr){2-4}
\cmidrule(lr){5-7}
\cmidrule(lr){8-10}
& CE & +~\our{} & $\Delta$ (\%)
& CE & +~\our{} & $\Delta$ (\%)
& CE & +~\our{} & $\Delta$ (\%) \\
\midrule[0.5pt]
fVLM
& 384.54 & 391.39 & $+1.82$
& 3.913 & 3.976 & $+1.62$
& 4.422 & 4.441 & $+0.44$ \\
BTB3D
& 434.88 & 453.96 & $+4.41$
& 56.770 & 56.775 & $+0.01$
& 58.375 & 60.055 & $+2.88$ \\
HLIP
& 414.97 & 413.24 & $-0.36$
& 3.420 & 3.420 & $0.00$
& 3.793 & 3.852 & $+1.54$ \\
RadFM
& 687.49 & 690.17 & $+0.43$
& 27.204 & 27.411 & $+0.76$
& 27.842 & 28.018 & $+0.63$ \\
\bottomrule[1pt]
\end{tabular*}
\end{table}

\begin{figure}[h]
    \centering
    \includegraphics[width=\linewidth]{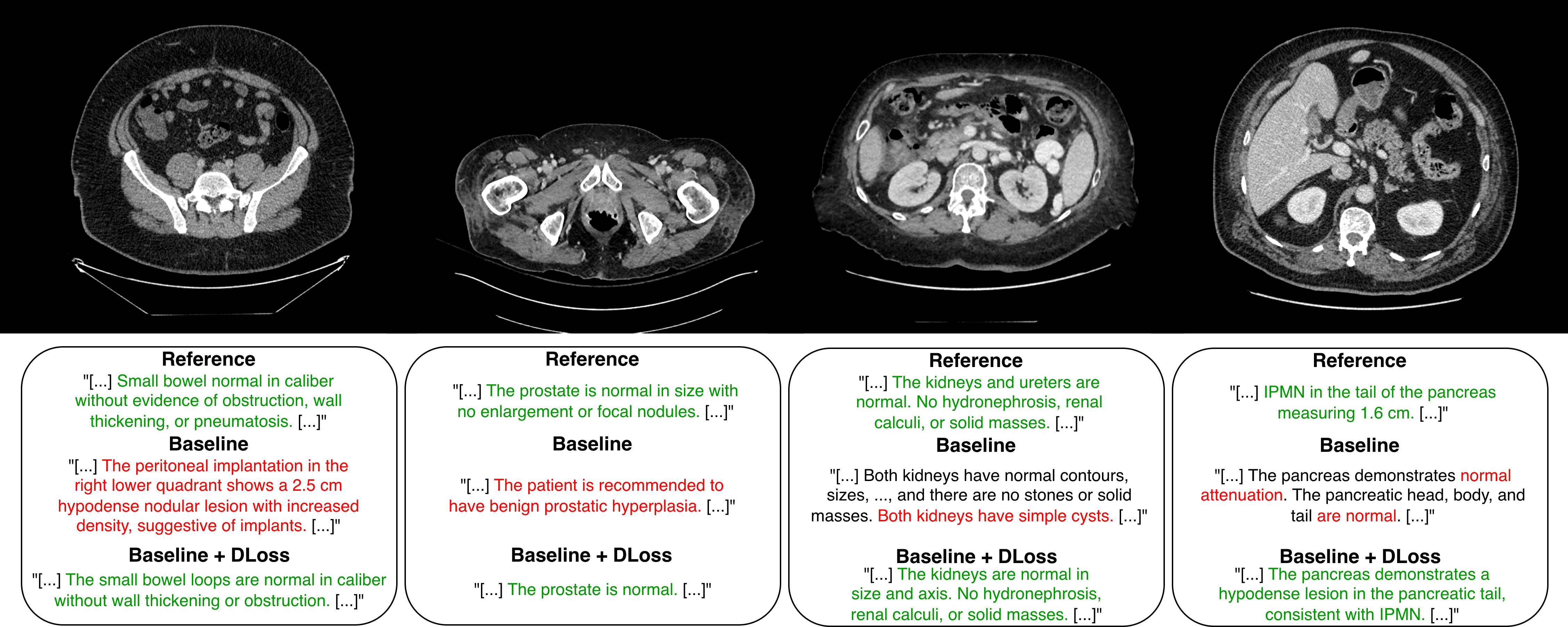}
    \caption{\textbf{Qualitative comparison of organ-level CT report generation.} Each column shows a representative CT slice and excerpts from the reference, baseline, and baseline + \our{} findings for one anatomical organ. \textcolor{green!55!black}{Green} text indicates content consistent with the reference; \textcolor{red!90!black}{red} text marks baseline statements not supported by the reference.}
    \label{fig:qualitativeresults}
\end{figure}

\noindent
\textbf{Qualitative results.} Figure~\ref{fig:qualitativeresults} compares organ-level report excerpts across four CT studies. The baseline model generates false positive findings that are absent in the ground truth references. Conversely, it can also produce false negatives by missing key clinical status updates. Incorporating \our{} addresses both issues by effectively mitigating false positives and false negatives, enabling accurate detection, classification, and reporting of diagnostic findings aligned with the reference standard. These examples show how \our{} enhances overall diagnostic faithfulness by eliminating unsupported findings while consistently retaining true positive and true negative findings.

\subsection{Ablation studies}

\begin{wraptable}{r}{0.45\textwidth}
\vspace{-4.0em}
\centering
\caption{\textbf{Component and pooling ablations of \our{} on fVLM.}
S/U denote supported/unsupported targets. ``Only'' retains CE and the indicated
DLoss component; ``max'' and ``mean'' replace that target's pooling operator.
The CE row gives absolute scores; other rows give relative percentage changes
from CE. }
\label{tab:diagnostic_loss_ablations}
\vspace{0.3em}
\small
\setlength{\tabcolsep}{2pt}
\resizebox{\linewidth}{!}{%
\begin{tabular}{lrrr}
\toprule
Config & BLEU-4 (\%) & GREEN (\%) & F1-RG (\%) \\
\midrule
\#1 CE & 49.55 & 65.96 & 55.55 \\
\#2 S-only & \textcolor{red!90!black}{$-1.88$} & \textcolor{red!90!black}{$-1.11$} & \textcolor{red!90!black}{$-0.74$} \\
\#3 U-only & \textcolor{red!90!black}{$-2.75$} & \textcolor{red!90!black}{$-2.62$} & \textcolor{red!90!black}{$-1.87$} \\
\#4 S-max & \textcolor{red!90!black}{$-16.11$} & \textcolor{red!90!black}{$-14.76$} & \textcolor{red!90!black}{$-11.10$} \\
\#5 S-mean & \textcolor{red!90!black}{$-2.27$} & \textcolor{red!90!black}{$-1.28$} & \textcolor{red!90!black}{$-1.91$} \\
\#6 U-mean & \textcolor{red!90!black}{$-22.85$} & \textcolor{red!90!black}{$-25.17$} & \textcolor{red!90!black}{$-16.37$} \\
\midrule
\#7 \textbf{\our{}} & \textcolor{green!55!black}{$\mathbf{+0.73}$} & \textcolor{green!55!black}{$\mathbf{+0.47}$} & \textcolor{green!55!black}{$\mathbf{+1.54}$} \\
\bottomrule
\end{tabular}%
}
\vspace{-1.5em}
\end{wraptable}

Table~\ref{tab:diagnostic_loss_ablations} evaluates component and pooling variants of \our{} on fVLM.
These ablations use random seed 13, running on the full Merlin test set against their matching CE baseline.
At \((\lambda_+,\lambda_-)=(0.001,0.00025)\), the full \our{} objective improves BLEU-4, GREEN, and F1-RG over the matched CE baseline.
Using only the supported (S-only) or unsupported (U-only) target loss degrades performance across all metrics, supporting the complementary contributions of both components in this setting.
Replacing supported-target noisy-OR with maximum (S-max) or mean (S-mean), or unsupported-target log-sum-exp with mean (U-mean), also lowers all three scores. U-mean produces the largest declines. Further details are in Appendix~\ref{sec:appendix_ablation_studies}.

\section{Conclusions}

In this work, we present DLoss, a plug-and-play 3D model-agnostic loss function based on the asymmetric pooling of token probabilities across decoder positions for radiology report generation. Extensive experiments across both in-distribution and out-of-distribution benchmarks show that integrating DLoss consistently enhances diagnostic faithfulness while maintaining competitive language-generation performance. Overall, this lightweight objective improves the performance of six state-of-the-art report generation methods across six evaluation metrics.

\noindent
\textbf{Limitations.}
The controlled diagnostic target space is derived from training reports, so
the auxiliary loss directly supervises only the concepts and expressions
represented there. Our experiments span six backbones and two test sets but
remain within abdominal CT, and the reported scores rely on automatic metrics. Testing \our{}
on 2D X-ray reports and comparing its outputs with radiologist judgments
would clarify how broadly the observed gains translate.

\section*{Acknowledgments}

This research was funded in part by National Science Center, Poland 2025/57/B/ST6/03100. We gratefully acknowledge Polish high-performance computing infrastructure PLGrid (HPC Center: ACK Cyfronet AGH) for providing computer facilities and support within computational grant no. PLG/2025/018792.

\subsection*{AI use statement}

In this work, we used generative AI tools to assist with refining manuscript
wording and drafting code. The authors designed the study, conducted the
experiments, and determined the interpretation of the results. References were
selected by the authors, and their bibliographic entries were copied manually
from source records. The authors reviewed AI-assisted material before
incorporating it and take full responsibility for the methods, results, and
content of this paper.

\subsection*{Ethics statement}

In this work, we used a private dataset for external evaluation. This study was approved by our Institutional Review Board (IRB). Approval of the Bioethics Committee at the Faculty of Medicine, Collegium Medicum of the University of Warmia and Mazury in Olsztyn, No. 9/2023, dated March 30, 2023

\subsection*{Reproducibility statement}

The diagnostic objective and training algorithm are specified in
Section~\ref{sec:diagnostic-loss-formulation}. The appendix documents dataset
composition and splits (Appendix~\ref{sec:appendixdata}), report preprocessing
and target construction (Appendices~\ref{sec:appendix_report_prompts} and
\ref{sec:appendix_diagnostic_supervision}), model and training details
(Appendix~\ref{sec:appendix_implementation_details}), evaluation metrics
(Appendix~\ref{sec:appendix_evaluation_metrics}), and the paired-seed analysis
(Appendix~\ref{sec:appendix_nine_seed_intervals}). The code for DLoss is available at \url{https://github.com/gmum/DLoss}. The in-house test cohort is not publicly available.



\bibliography{iclr2027_conference}
\bibliographystyle{iclr2027_conference}

\newpage

\appendix

\renewcommand \thepart{}
\renewcommand \partname{}

\part{Appendix} 
\setcounter{secnumdepth}{4}
\setcounter{tocdepth}{4}
\parttoc 

\clearpage

\section{Data}
\label{sec:appendixdata}

We evaluate Diagnostic Loss on abdominal CT data. Table~\ref{tab:3ddataappendix} summarizes the datasets used for training and testing. Section~\ref{sec:trainingdataappendix} describes the training data, and Section~\ref{sec:testdataappendix} describes the test data.

\begin{table*}[ht]
    \centering
    \caption{\textbf{Overview of the 3D datasets used for training or testing.} We provide the number of scans, exams, patients, and training and test splits for each dataset.}
    \label{tab:3ddataappendix}
    \begin{tabular}{lccccc} 
        \toprule
        \textbf{Dataset} & \textbf{Scans} & \textbf{Exams} & \textbf{Patients} & \textbf{Train} & \textbf{Test}\\ 
        \midrule
        Merlin  & 25,494  & 25,494 & 18,317 & 20,369  & 5,125\\
        CancerVerse & 24,168 & 22,867 & 13,863 & 24,168  & - \\
        In-house & 2,706 & 2,706 & 1,948 & - & 2,706 \\
        \midrule
        \textbf{Total}    & \textbf{52,368}  & \textbf{50,767}  & \textbf{34,128}  & \textbf{44,537} & \textbf{7,831}  \\
        \bottomrule
    \end{tabular}
\end{table*}

\subsection{Training set}
\label{sec:trainingdataappendix}

For the 3D experiments reported here, we train models with and without Diagnostic Loss using Merlin~\citep{blankemeier2026merlin} and CancerVerse~\citep{li2026cancerverse}.

\subsubsection{Merlin} The training set consists of 20,369 contrast-enhanced CT scans. Following \citet{blankemeier2026merlin}, we split the dataset for training (15,309 CTs), and validation (5,060 CTs) sets. Each CT exam is paired with the corresponding radiology report. The
dataset includes abdominal and pelvis CT exams conducted between 2012 and 2018 at the Stanford Hospital Emergency Departments. 

\subsubsection{CancerVerse} 
The training set consists of 24,168 multi-phase CT scans from 13,863 patients\footnote{CancerVerse dataset is publicly available at \url{https://huggingface.co/datasets/BodyMaps/CancerVerse}}. We use all cases as a training set. Each CT exam is paired with the corresponding radiology report. The dataset includes abdominal and pelvis CT scans acquired at the University Hospital Basel, Switzerland and Medipol University Hospital, Türkiye.


\subsection{Test set}
\label{sec:testdataappendix}

\subsubsection{Merlin} 
We use an official test set \citep{blankemeier2026merlin}, which consists of 5,125 contrast-enhanced CT scans.

\subsubsection{In-house} 
The dataset consists of 2,706 multi-center, contrast-enhanced CT scans from 1,948 patients. Each CT examination is paired with the corresponding radiology report. All scans were acquired across 300 sites within a nationwide cohort. Approval of the Bioethics Committee at the Faculty of Medicine, Collegium Medicum of the University of Warmia and Mazury in Olsztyn, No. 9/2023, dated March 30, 2023



\providecolor{jsonteal}{HTML}{0F766E}

\section{Report preprocessing}
\label{sec:appendix_report_prompts}

\begin{figure*}[h]
\centering

\begin{tcolorbox}[
  width=\textwidth,
  colframe=reportblue!80,
  colback=reportblue!4,
  coltitle=white,
  fonttitle=\bfseries\footnotesize,
  title=Free-text CT report,
  colbacktitle=reportblue!80,
  fontupper=\small,
]
\textit{FINDINGS:} Lower thorax: Mild atelectatic change bilaterally. Liver and biliary tree: Nonspecific perfusion changes. No suspicious hepatic lesions. Gallbladder: Gallstones without CT findings of cholecystitis. Spleen: Normal. Pancreas: Normal. Adrenal glands: Normal. Kidneys and ureters: Simple appearing renal cysts. Gastrointestinal tract: Diverticulosis without evidence of acute diverticulitis. Sequela of prior inflammation involving the rectosigmoid colon. Peritoneal cavity: No free fluid. Bladder: Normal. Prostate and seminal vesicles: Prostate is mildly prominent size measuring 5 cm in transverse dimension. Vasculature: Patent. No aneurysm. Lymph nodes: Normal. Abdominal wall: Normal. Musculoskeletal: Degenerative change of the spine.
\end{tcolorbox}

\begin{center}
\begin{tikzpicture}[>=Latex]
  \draw[very thick, -{Latex[length=4mm]}, qwenpurple]
    (0,0) -- (0,-1.4);
  \node[right=0.6cm] at (0,-0.7) {%
    \includegraphics[height=6em]{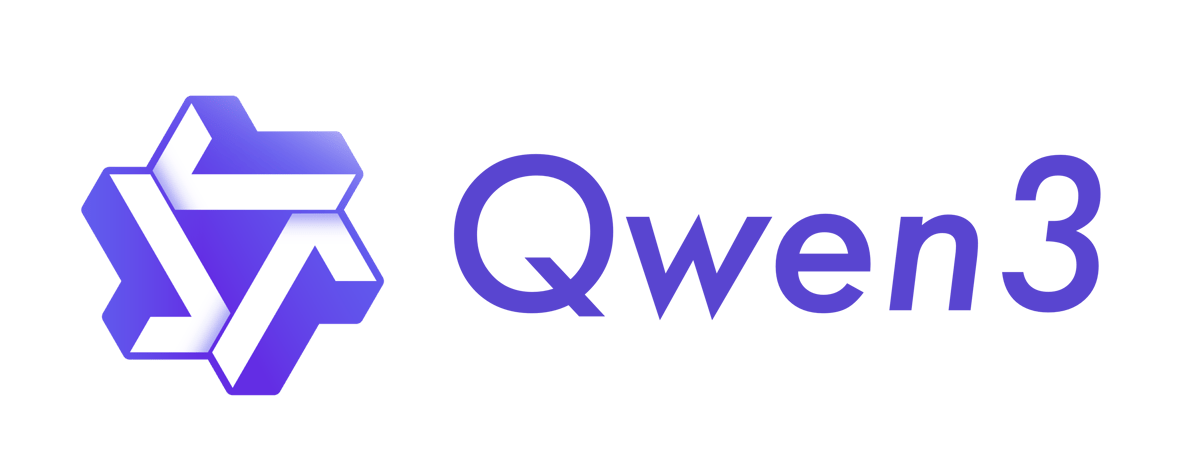}%
  };
\end{tikzpicture}
\end{center}

\begin{tcolorbox}[
  colframe=jsonteal!80,
  colback=jsonteal!4,
  coltitle=white,
  fonttitle=\bfseries\footnotesize,
  title=Structured output,
  colbacktitle=jsonteal!80,
  fontupper=\scriptsize,
  before skip=6pt,
  after skip=3pt
]
\begin{verbatim}
{
  "study_id": "AC423fdb2",
  "cleaned_report": "[... The full cleaned report text is here ...]",
  "organs": {
    "Spleen": "The spleen is normal in size and homogeneous, with no splenomegaly or
    focal lesions identified.",
    "Kidneys": "Simple appearing renal cysts.",
    "Gallbladder": "Gallstones without CT findings of cholecystitis.",
    "Liver": "Nonspecific perfusion changes. No suspicious hepatic lesions.",
    "Stomach": "Unremarkable",
    "Pancreas": "The pancreas demonstrates normal attenuation. No pancreatic duct
    dilatation or focal masses.",
    "Adrenal glands": "The adrenal glands have normal morphology with no nodules or
    masses.",
    "Small bowel": "Unremarkable",
    "Colon": "Diverticulosis without evidence of acute diverticulitis. Sequela of
    inflammation involving the rectosigmoid colon.",
    "Urinary bladder": "The urinary bladder is well-distended with no wall
    thickening or masses.",
    "Prostate": "The prostate is mildly prominent size measuring 5 cm in transverse
    dimension."
  }
}
\end{verbatim}
\end{tcolorbox}
\caption{\textbf{Example of report-to-structured-text conversion.} Qwen3 maps a free-text CT report to a cleaned report and findings for each of the 11 organ groups. The cleaned report is omitted for display.}
\label{fig:preprocessing}
\end{figure*}

Figure~\ref{fig:preprocessing} shows a free-text CT report and an abbreviated structured output from Qwen3.

\clearpage

\subsection{Report preprocessing example}
\label{sec:appendix_report_preprocessing}

\begin{figure*}[htbp]
    \centering
    \includegraphics[width=\linewidth]{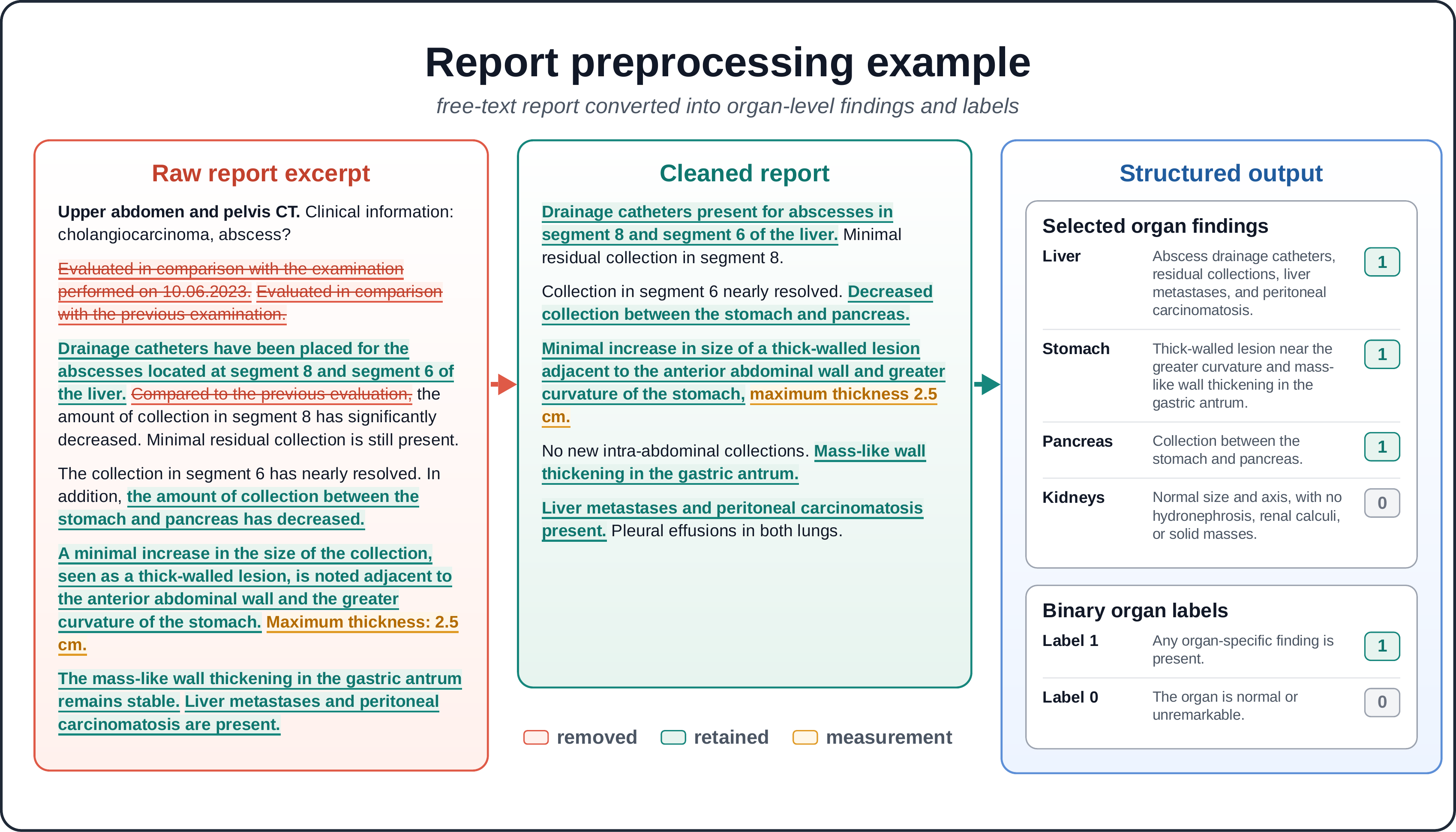}
    \caption{\textbf{Example of radiology report preprocessing for a single CT study.} The red panel shows the raw report excerpt, the green panel
    shows the cleaned current-study report text, and the blue panel shows selected
    organ-specific outputs and binary organ labels. Text highlights indicate
    removed phrasing, retained findings, and retained measurements. Only selected
    organs are shown for readability; the full preprocessing output contains all
    11 organ groups.}
    \label{fig:report_preprocessing_example}
\end{figure*}

Figure~\ref{fig:report_preprocessing_example} shows how one report is cleaned and separated into organ-specific texts and labels.

\clearpage
\section{Diagnostic supervision construction}
\label{sec:appendix_diagnostic_supervision}

\begin{figure*}[htbp]
    \centering
    \includegraphics[width=\linewidth]{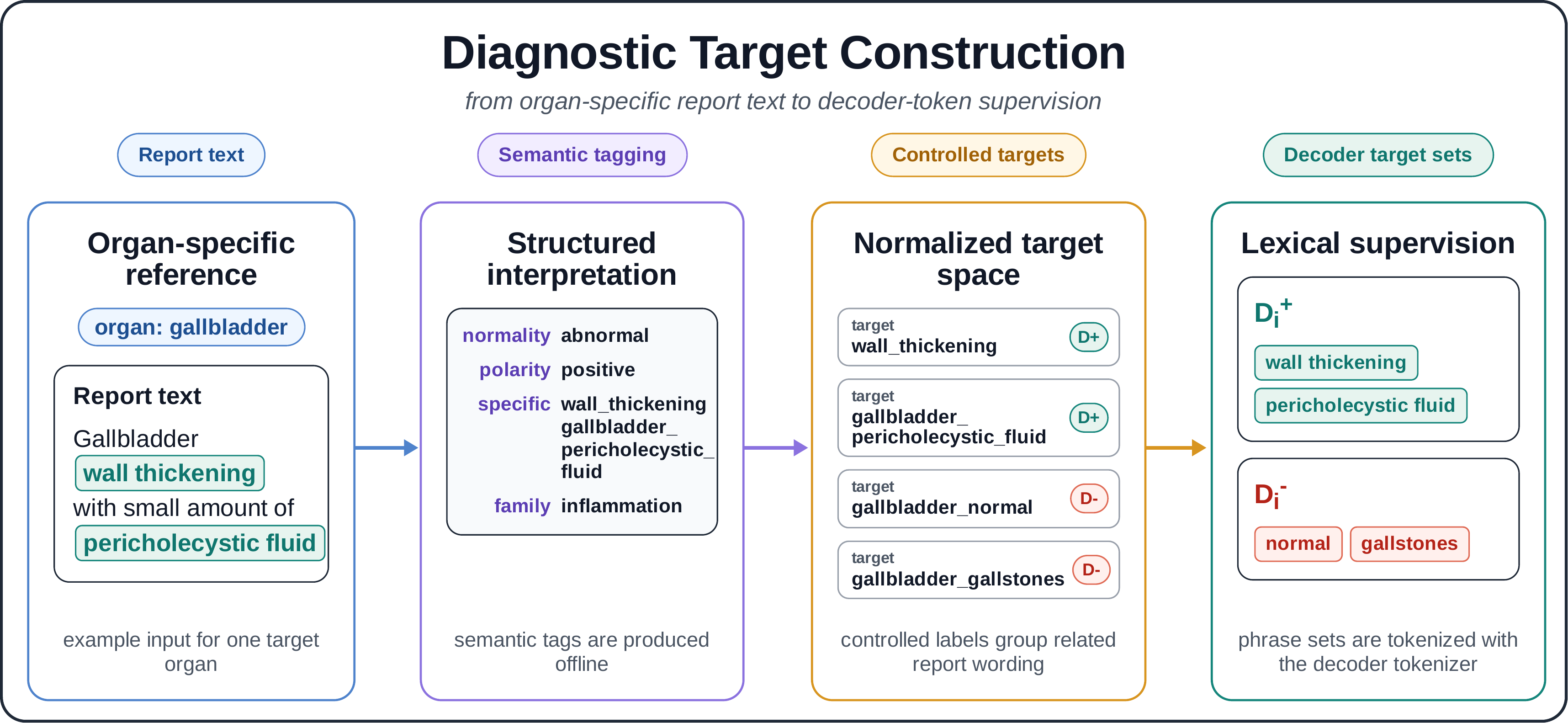}
    \caption{\textbf{Schematic overview of diagnostic target construction for the
    diagnostic loss.} An organ-specific reference text is assigned structured
    semantic tags, mapped into a controlled diagnostic target space, and
    converted into supported and unsupported lexical target sets for the decoder
    loss. Supported targets are findings stated in the reference text; unsupported
    targets are alternatives for the same organ that are discouraged during
    training. The gallbladder example displays representative targets and
    lexical forms for one organ.}
    \label{fig:diagnostic_supervision_construction}
\end{figure*}

Figure~\ref{fig:diagnostic_supervision_construction} illustrates how supported and unsupported diagnostic targets are constructed from an organ-specific reference text.

\clearpage
\section{Diagnostic loss illustrations}
\label{sec:appendix_diagnostic_loss_details}

\begin{figure*}[htbp]
    \centering
    \includegraphics[width=\linewidth]{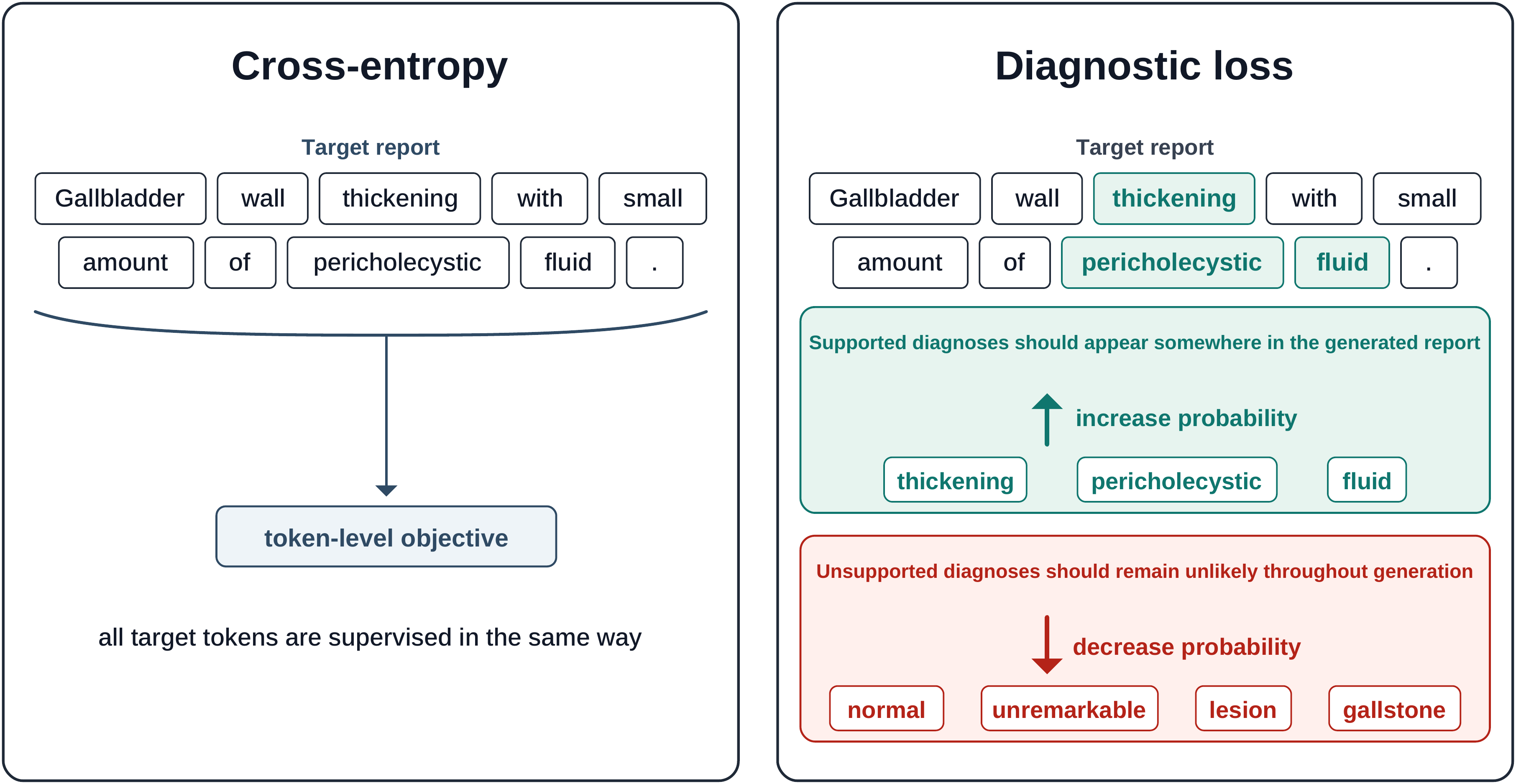}
    \caption{\textbf{Conceptual comparison between token-level cross-entropy and the
    proposed \emph{diagnostic loss}.} Cross-entropy supervises each next-token prediction in the
    reference radiology report. Diagnostic loss increases probability assigned to
    supported targets at one or more decoder positions and suppresses probability
    assigned to unsupported targets across positions. For readability, tokens are shown as whole words, although model
    tokenizers may split words into smaller units.}
    \label{fig:cross_entropy_vs_diagnostic_loss}
\end{figure*}

Figure~\ref{fig:cross_entropy_vs_diagnostic_loss} contrasts the two training objectives.

\clearpage

\section{Experimental setup}
\label{sec:experimental_setup}

\subsection{Implementation details}
\label{sec:appendix_implementation_details}

We implement our method using PyTorch and train on up to four NVIDIA GH200 GPUs. During report-generation training, visual features are fixed. We train low-rank adaptation (LoRA) parameters in each language decoder together with its visual-to-text projector or adapter, using the same visual initialization for the cross-entropy and Diagnostic Loss runs of each backbone. Merlin and RadFM use released visual weights, SPECTRE uses a pretrained visual checkpoint, and fVLM, BTB3D, and HLIP use locally trained visual checkpoints.

Merlin uses the RadLLaMA-7B decoder; BTB3D uses Llama-3.1-8B-Instruct; fVLM, HLIP, and SPECTRE use Qwen2.5-0.5B; and RadFM uses MedLLaMA-13B. Each report generator predicts one organ finding at a time from an organ prompt. For each backbone, the cross-entropy and Diagnostic Loss runs use the same decoder, tokenizer, visual initialization, and pooled Merlin--CancerVerse report-generation training data.

The full decoder weights remain frozen. Merlin uses LoRA rank 512, alpha 16, and dropout 0.1 on the decoder attention query and value projections. BTB3D uses rank 128, alpha 256, and dropout 0.05; fVLM, HLIP, RadFM, and SPECTRE use rank 16, alpha 32, and dropout 0.05. These latter five models apply LoRA to the decoder query, key, value, and output projections and to the gate, up, and down projections of its feed-forward layers. For each backbone, the cross-entropy and Diagnostic Loss runs use the same LoRA configuration and trainable projector or adapter.

\noindent\textbf{Diagnostic Loss settings.}
The backbone-specific loss coefficients in the main experiments were selected
through a small grid search and are listed in Table~\ref{tab:main_dloss_settings}.
All runs use $\tau=8$.

\begin{table}[ht]
\centering
\caption{\textbf{Diagnostic Loss coefficients in the main experiments.}
The columns correspond to the supported- and unsupported-target coefficients
in Eq.~\ref{eq:sample-diagnostic-loss}.}
\label{tab:main_dloss_settings}
\small
\begin{tabular}{lcc}
\toprule
Backbone & $\lambda_+$ & $\lambda_-$ \\
\midrule
Merlin--RadLLaMA & 0.001 & 0.0005 \\
fVLM--Qwen & 0.001 & 0.00025 \\
BTB3D & 0.001 & 0.00025 \\
HLIP--Qwen & 0.001 & 0.00025 \\
RadFM-v3 & 0.01 & 0.01 \\
SPECTRE--Qwen & 0.01 & 0.005 \\
\bottomrule
\end{tabular}
\end{table}

\subsection{Evaluation metrics}
\label{sec:appendix_evaluation_metrics}

\subsubsection{Lexical metrics}
The lexical metrics used in the experiments compare generated and reference
texts through shared tokens, ordered word sequences, or weighted phrase overlap.
Most of them operate on \emph{n}-grams, meaning contiguous groups of \(n\)
tokens: unigrams contain one token, bigrams contain two adjacent tokens, and
longer \emph{n}-grams capture longer local phrases.
They are computed using the Microsoft Common Objects in Context (MS COCO)
caption-evaluation setup
\citep{chen2015microsoft}.
The equations below state the standard nondegenerate definitions. In the
implementation, zero modified precisions and empty generated texts are handled
by the small constants and brevity penalty used in the COCO caption-evaluation
scorer, rather than by passing zero directly to the logarithm. For METEOR,
examples with no matched unigrams receive zero score, so the chunk penalty is
only relevant when at least one unigram is matched. For CIDEr, cosine-similarity
terms with a zero-norm TF-IDF vector are treated as zero.

\textbf{BLEU-4}~\citep{papineni2002bleu} measures how many generated
\emph{n}-grams also appear in the reference. Plain precision would allow a
model to repeat a correct word many times and receive excessive credit, so BLEU
uses clipped counts: each generated \emph{n}-gram can be counted only up to its
frequency in the reference. For a generated text \(c_i\) and a reference text
\(r_i\), the modified precision for order \(n\) is:
\begin{equation}
    p_n =
    \frac{
        \sum_{g \in G_n(c_i)}
        \min\left(\mathrm{count}_{c_i}(g), \mathrm{count}_{r_i}(g)\right)
    }{
        \sum_{g \in G_n(c_i)} \mathrm{count}_{c_i}(g)
    } .
\end{equation}
BLEU-4 combines the modified precisions for \(n=1,\ldots,4\) using a geometric
mean and applies a brevity penalty:
\begin{equation}
    \mathrm{BLEU}_4 =
    \mathrm{BP}
    \exp\left(\frac{1}{4}\sum_{n=1}^{4}\log p_n\right),
\end{equation}
\begin{equation}    
\mathrm{BP} =
    \begin{cases}
        1, & |c_i| > |r_i|, \\
        \exp\left(1 - \frac{|r_i|}{|c_i|}\right), & |c_i| \le |r_i|.
    \end{cases}
\end{equation}
Here \(|c_i|\) and \(|r_i|\) denote text lengths after metric tokenization.
The brevity penalty reduces the score when the generated text is shorter than
the reference, because high precision alone can otherwise be obtained by
generating only a small subset of the radiology report. BLEU is therefore sensitive to how
much of the generated wording is supported by the reference, but it is less
direct about missing reference content.

\textbf{ROUGE-L}~\citep{lin2004rouge} measures overlap through the longest
common subsequence (LCS) between the generated and reference texts. A
subsequence preserves word order but does not require the matched words to be
adjacent. If \(L_i\) is the LCS length, ROUGE-L defines an LCS-based precision
and recall:
\begin{equation}
    P_L = \frac{L_i}{|c_i|}, \qquad
    R_L = \frac{L_i}{|r_i|}.
\end{equation}
These quantities are combined with an F-measure:
\begin{equation}
    \mathrm{ROUGE\text{-}L} =
    \frac{(1+\beta^2)P_LR_L}{R_L + \beta^2P_L}.
\end{equation}
The parameter \(\beta\) controls the precision--recall trade-off. This makes
ROUGE-L more sensitive to reference coverage than BLEU, because it rewards
generated texts that preserve longer ordered sequences from the reference.

\textbf{METEOR}~\citep{lavie2009meteor} starts from unigram alignment rather
than higher-order \emph{n}-gram matching. The generated and reference texts are
aligned using exact matches, stems, and synonyms, so two wording variants can
receive credit even when their surface forms are not identical. If \(m_i\)
tokens are matched, the METEOR precision, recall, and weighted harmonic mean are:
\begin{equation}
    P_M = \frac{m_i}{|c_i|}, \qquad
    R_M = \frac{m_i}{|r_i|}, \qquad
    F_{\mathrm{mean}} = \frac{10P_MR_M}{R_M + 9P_M}.
\end{equation}
This weighted harmonic mean gives more weight to recall than to precision.
METEOR then checks whether the matched words form long contiguous runs or are
split across the sentence. Let \(q_i\) be the number of matched chunks and
let \(m_i\) be the number of matched unigrams. The penalty has the form:
\begin{equation}
    \mathrm{Penalty} =
    \gamma\left(\frac{q_i}{m_i}\right)^{\theta},
\end{equation}
where \(\gamma\) and \(\theta\) are fixed metric parameters from the METEOR
implementation. The final score is:
\begin{equation}
    \mathrm{METEOR} = (1 - \mathrm{Penalty})F_{\mathrm{mean}}.
\end{equation}
This gives the metric a different behavior from pure overlap counts, because
some wording variants can receive credit when they are aligned by the matching
modules.

\textbf{CIDEr}~\citep{vedantam2015cider} also compares \emph{n}-gram overlap,
but it weights \emph{n}-grams by how informative they are in the evaluation
collection. Common phrases receive less weight, while rarer phrases receive
more weight. For an \emph{n}-gram \(g\), the term frequency in text \(x\) is:
\begin{equation}
    \mathrm{tf}_{n,g}(x) =
    \frac{\mathrm{count}_x(g)}
    {\sum_{g' \in G_n(x)} \mathrm{count}_x(g')}.
\end{equation}
Let \(\mathrm{df}_{n,g}\) be the number of reference texts in the evaluation
set that contain \(g\). The inverse document frequency is:
\begin{equation}
    \mathrm{idf}_{n,g} =
    \log\left(\frac{N}{\mathrm{df}_{n,g}}\right),
\end{equation}
with standard smoothing in the implementation to avoid undefined values. The
term-frequency--inverse-document-frequency (TF-IDF) vector for order \(n\) has components:
\begin{equation}
    v_{n,g}(x) =
    \mathrm{tf}_{n,g}(x)\mathrm{idf}_{n,g}.
\end{equation}
Let \(\bm{v}_n(c_i)\) and \(\bm{v}_n(r_i)\) be the resulting vectors for the
generated and reference texts. The score for one \emph{n}-gram order is their
cosine similarity:
\begin{equation}
    \mathrm{CIDEr}_n(c_i,r_i) =
    \frac{\bm{v}_n(c_i) \cdot \bm{v}_n(r_i)}
    {\left\|\bm{v}_n(c_i)\right\|\left\|\bm{v}_n(r_i)\right\|}.
\end{equation}
The final CIDEr score averages the first four \emph{n}-gram orders:
\begin{equation}
    \mathrm{CIDEr}(c_i,r_i) =
    \frac{10}{4}\sum_{n=1}^{4}\mathrm{CIDEr}_n(c_i,r_i).
\end{equation}
The factor \(10\) is the conventional CIDEr scaling used by the metric
definition. Compared with unweighted overlap, CIDEr gives more influence to
phrases that are informative in the evaluation collection.

\subsubsection{Clinical-content metrics}

Clinical-content metrics make the distinction between lexical and clinical evaluation operational by comparing radiology report meaning instead of only surface wording. This matters for diagnostic loss because the
loss is designed to affect diagnostic expressions derived from the reference
report in the decoder distribution.
The experiments compute RadGraph and GREEN, or Generative Radiology Report
Evaluation and Error Notation~\citep{ostmeier2024green}, with the RadEval
framework \citep{xu2025radeval}.
These metrics compare generated and reference reports automatically; they are
not a radiologist review tailored to individual CT organs and do not establish
clinical safety on their own.

\textbf{RadGraph}~\citep{jain1radgraph} evaluates radiology reports by applying an
automatic clinical information-extraction model to both the generated and
reference texts. The RadGraph paper defines an annotation schema for radiology
reports and trains a benchmark extractor to predict this schema from free text.
The extractor first identifies text spans that correspond to anatomy and
observations, then assigns observation status, such as definitely present,
uncertain, or definitely absent, and predicts relations between entity pairs.
The benchmark model reported by RadGraph uses a transformer-based entity and
relation extraction architecture with biomedical language-model initialization.
A generated radiology report therefore receives credit only when the extracted clinical
structure agrees with the reference: what observation is mentioned, where it is
located, and whether it is asserted or negated. RadEval returns simple,
partial, and complete RadGraph variants; the experiments report the complete
entity-and-relation score as \(\mathrm{F1}_{\mathrm{RG}}\).
If \(T(c_i)\) and \(T(r_i)\) denote the entity-and-relation tuples extracted
from the generated and reference texts, RadGraph precision and recall are:
\begin{equation}
    P_{\mathrm{RG}} =
    \frac{|T(c_i) \cap T(r_i)|}{|T(c_i)|}, \qquad
    R_{\mathrm{RG}} =
    \frac{|T(c_i) \cap T(r_i)|}{|T(r_i)|},
\end{equation}
where the intersection counts extracted tuples that match between the generated
and reference radiology reports. The complete RadGraph F1 score is:
\begin{equation}
    \mathrm{F1}_{\mathrm{RG}} =
    \frac{2P_{\mathrm{RG}}R_{\mathrm{RG}}}{P_{\mathrm{RG}}+R_{\mathrm{RG}}}.
\end{equation}
If either report is empty or either extracted graph has no entities, the
corresponding RadGraph score contribution is treated as zero by the RadEval
implementation.
The reported value is the mean \(\mathrm{F1}_{\mathrm{RG}}\) over all evaluated
organ-specific examples.

\textbf{GREEN} uses a generative language model as the evaluator. The
evaluator receives a reference radiology report and a candidate radiology report, then produces a
structured error analysis rather than only a single similarity value. The
original GREEN method uses GPT-4 analyzes of synthetic radiology report pairs to train
radiology-domain language models that identify and classify report errors.
These errors include false findings, missing findings, wrong anatomical
location, wrong severity, and unsupported comparison statements. GREEN also
distinguishes clinically significant errors from less important discrepancies,
which makes the score closer to an error-based clinical review than to lexical
overlap. In the experiments, RadEval computes GREEN with the released local
GREEN-RadLLaMA2-7b evaluator rather than
with an API-based GPT-4 judge.

The scalar GREEN score is the aggregation of the structured evaluator output
used in the evaluation implementation. Let \(M_i\) be the number of matched
findings in example \(i\), and let
\(E_{i,k}^{\mathrm{sig}}\) be the number of clinically significant errors in
error category \(k\). The per-example score is:
\begin{equation}
    s_i^{\mathrm{GREEN}} =
    \begin{cases}
        \dfrac{M_i}{M_i + \sum_k E_{i,k}^{\mathrm{sig}}},
        & M_i > 0, \\
        0, & M_i = 0.
    \end{cases}
\end{equation}
This expression summarizes the structured GREEN error output as a scalar score:
matched findings increase the numerator, while clinically significant errors
increase the denominator. Higher values, therefore, indicate that the generated
radiology report preserves more reference findings while introducing fewer
clinically significant errors. The reported GREEN value is the mean over the
evaluated examples:
\begin{equation}
    \mathrm{GREEN} =
    \frac{1}{N}\sum_{i=1}^{N}s_i^{\mathrm{GREEN}} .
\end{equation}

\clearpage
\section{Robustness across training seeds}
\label{sec:appendix_nine_seed_intervals}

Table~\ref{tab:3dradiologyreportsperformance} shows gains from Diagnostic Loss across several backbones, with one training run per setting. Repeating every backbone across many seeds would require substantial additional training. We therefore complement the cross-backbone comparison with nine matched BTB3D runs on Merlin. This experiment tests whether the BTB3D gains persist across training seeds and resampled test studies.

We compared BTB3D trained with cross-entropy (CE) and with CE plus Diagnostic Loss (DLoss) using nine matched random seeds: 7, 13, 42, 101, 103, 107, 2026, 2718, and 31415. Each model was evaluated on the full Merlin test set of 5,125 studies and 56,375 organ-specific report rows. CE and DLoss runs with the same seed use the same test rows. All rows, including empty generations, remain in the metric denominators.

\noindent\textbf{Statistical analysis.} For each metric, we subtract the CE score from the DLoss score within each matched seed pair, then average the nine differences. A positive mean effect favors DLoss. Table~\ref{tab:paired_seed_overall} reports this effect and two bootstrap 95\% confidence intervals, each based on 5,000 resamples. The seed bootstrap samples the nine CE--DLoss seed pairs with replacement and recomputes the mean effect. The seed--study bootstrap also samples test studies with replacement. It retains all 11 organ rows from each sampled study and uses the same sampled studies for CE and DLoss in every selected seed pair. Each interval spans the 2.5th to 97.5th percentiles of its resampled effects. Pairing preserves the comparison within a training seed; sampling by study respects the clustering of organ rows. For corpus BLEU, we recompute the score from $n$-gram counts in each resample. Other metrics use saved row scores, with the original test-corpus inverse document frequencies held fixed for CIDEr. The tables report effects on the native scales of the saved metric outputs.

\begin{table}[htbp]
\centering
\caption{\textbf{Paired BTB3D results on the full Merlin test set.} Each effect is the mean DLoss-minus-CE difference over nine matched seeds. Intervals are 95\% confidence intervals; positive values favor DLoss. Point estimates use all 56,375 organ rows per run.}
\label{tab:paired_seed_overall}
\small
\setlength{\tabcolsep}{4pt}
\begin{tabular}{lccc}
\toprule
Metric & Mean effect & Seed bootstrap & Seed--study bootstrap \\
\midrule
BLEU-1 & $+0.0616$ & $[+0.0046, +0.1242]$ & $[+0.0050, +0.1238]$ \\
BLEU-2 & $+0.0698$ & $[+0.0099, +0.1387]$ & $[+0.0096, +0.1385]$ \\
BLEU-3 & $+0.0738$ & $[+0.0114, +0.1460]$ & $[+0.0116, +0.1450]$ \\
BLEU-4 & $+0.0752$ & $[+0.0124, +0.1483]$ & $[+0.0127, +0.1475]$ \\
METEOR & $+0.0795$ & $[+0.0169, +0.1480]$ & $[+0.0164, +0.1483]$ \\
ROUGE-L & $+0.0689$ & $[+0.0138, +0.1275]$ & $[+0.0137, +0.1277]$ \\
CIDEr & $+0.7834$ & $[+0.1528, +1.4527]$ & $[+0.1539, +1.4560]$ \\
GREEN & $+0.0455$ & $[+0.0197, +0.0728]$ & $[+0.0197, +0.0732]$ \\
RadGraph Complete & $+0.0795$ & $[+0.0203, +0.1447]$ & $[+0.0203, +0.1451]$ \\
\bottomrule
\end{tabular}
\end{table}

GREEN and RadGraph Complete have positive mean differences, and both reported bootstrap intervals lie above zero. The seven lexical metrics have positive mean differences and positive bootstrap intervals as well. Table~\ref{tab:paired_seed_individual} shows variation across runs: GREEN and RadGraph Complete improve in eight of nine seed pairs, while BLEU-4 improves in seven of nine. The positive mean effects do not imply improvement for every trained model.

\clearpage
\subsection{Variation across individual runs}

Table~\ref{tab:paired_seed_individual} shows the within-seed differences for BLEU-4 and the two clinical-content metrics. It makes the spread across runs visible alongside the average effects and intervals above.

\begin{table}[H]
\centering
\caption{\textbf{Within-seed effects for representative metrics.} Each entry is the DLoss-minus-CE difference on the full Merlin test set, in the native metric scale. These are paired training runs, not independent test-row samples.}
\label{tab:paired_seed_individual}
\small
\begin{tabular}{rrrr}
\toprule
Seed & BLEU-4 & GREEN & RadGraph Complete \\
\midrule
7 & $+0.0617$ & $+0.0726$ & $+0.0371$ \\
13 & $+0.2790$ & $+0.0913$ & $+0.2497$ \\
42 & $+0.0267$ & $+0.0433$ & $+0.0689$ \\
101 & $+0.0861$ & $+0.0205$ & $+0.0918$ \\
103 & $+0.0045$ & $+0.0141$ & $+0.0165$ \\
107 & $+0.2391$ & $+0.1216$ & $+0.2406$ \\
2026 & $-0.0449$ & $+0.0307$ & $-0.0611$ \\
2718 & $-0.0225$ & $-0.0215$ & $+0.0184$ \\
31415 & $+0.0472$ & $+0.0372$ & $+0.0538$ \\
\bottomrule
\end{tabular}
\end{table}

The bootstrap intervals are unadjusted for multiple metrics. With nine seed pairs, they describe uncertainty for this BTB3D protocol on Merlin; they do not establish seed robustness for the other backbones or the external test cohort.

\section{Ablation study details}
\label{sec:appendix_ablation_studies}

Table~\ref{tab:diagnostic_loss_ablations} compares matched fVLM runs with seed
13 on the full Merlin test set of 5,125 studies and 56,375 organ findings.
All percentage changes use the cross-entropy (CE) run from this ablation as
their baseline. Checkpoints are selected by validation CE. S-only adds the
supported-target loss to CE; U-only adds the unsupported-target loss to CE.
The pooling variants add both diagnostic components to CE and change one
pooling operator.
The ablations used two NVIDIA A100 GPUs (40 GB each), whereas the main
experiments used NVIDIA GH200 GPUs.

The S-max and S-mean variants replace supported-target noisy-OR in
Eq.~\ref{eq:positive-occurrence-probability} with
\begin{equation}
q_{i,d}^{\mathrm{S\mbox{-}max}}=\max_{t\in\mathcal{P}_i}m_{i,d,t},
\qquad
q_{i,d}^{\mathrm{S\mbox{-}mean}}=\frac{1}{|\mathcal{P}_i|}
\sum_{t\in\mathcal{P}_i}m_{i,d,t}.
\end{equation}
Both use $-\log q_{i,d}$ as in Eq.~\ref{eq:positive-target-loss}.
U-mean instead replaces Eq.~\ref{eq:negative-target-loss} with
\begin{equation}
L_{i,d}^{-,\mathrm{U\mbox{-}mean}}
=\log\left(\frac{1}{|\mathcal{P}_i|}
\sum_{t\in\mathcal{P}_i}m_{i,d,t}\right).
\end{equation}
Thus, the logarithm is applied after averaging the unsupported-target
probability masses. This variant does not use $\tau$; supported-target pooling
remains noisy-OR. As in the main equations, probability clamping is omitted
for readability.

Full \our{} improves all three reported metrics over the matched CE run,
whereas either component alone lowers them. For supported targets, maximum
pooling discards evidence outside the strongest position, while a mean can
dilute evidence localized to a few positions; both alternatives score below
noisy-OR here. U-mean has the largest declines, consistent with the need to
penalize high unsupported-target probability even when it occurs at a single
position. This run also generates 1,423 empty findings (2.52\%), compared with
48 (0.09\%) for CE; all are included in the metric denominators. Together,
these fVLM ablations support the two-component objective and its asymmetric
pooling rules.

\end{document}